\documentclass[
]{ceurart}
\begin{document}

\copyrightyear{2020}
\copyrightclause{Copyright for this paper by its authors.
  Use permitted under Creative Commons License Attribution 4.0
  International (CC BY 4.0).}

\conference{Woodstock'20: Symposium on the irreproducible science,
  June 01--05, 2020, Woodstock, NY}

\title{NLP is Not enough - Contextualization of User Input in Chatbots}

\author[1]{Nathan Dolbir}
\author[2]{Triyasha Dastidar}
\author[1]{Kaushik Roy}
\address[1]{Artificial Intelligence Institute, University of South Carolina}
\address[2]{BITS-Pilani Hyderabad}



\begin{abstract}
 AI chatbots have made vast strides in technology improvement in recent years and are already operational in many industries. Advanced Natural Language Processing techniques, based on deep networks, efficiently process user requests to carry out their functions. As chatbots gain traction, their applicability in healthcare is an attractive proposition due to the reduced economic and people costs of an overburdened system. However, healthcare bots require safe and medically accurate information capture, which deep networks aren't yet capable of due to user text and speech variations. Knowledge in symbolic structures is more suited for accurate reasoning but cannot handle natural language processing directly. Thus, in this paper, we study the effects of combining knowledge and neural representations on chatbot safety, accuracy, and understanding. 
\end{abstract}


\maketitle
\section{Introduction}
Conversational AI in its brief existence has already been extensively deployed in nearly every economic sector and is becoming increasingly easier for businesses to use. Its ability to mirror human conversation has made it a cost-effective solution to many customer service issues and it can provide users a stream of knowledge bounded only by the information it can access. The issue with AI's ability to mirror human natural language is that it doesn't have any true understanding 
of the language it is processing, but it can recognize patterns and characteristics of text based on repeated feeding of test input and by classification of a word and sentence meanings into a user-designed schema. Pattern recognition has evolved into deep learning 
of patterns it recognized in the context of the training a user has performed. Deep learning models have excelled at Natural Language Processing (NLP) tasks such as question-answering from the text in recent years \cite{otter2020survey}. Their widespread success has led to speedy adoption in real-world industrial applications \cite{li2018deep}. However, these models depend on pattern recognition and prediction techniques which often fail to correctly contextualize bits of input, especially with open-ended inputs based more on natural speech than an easily processed formulaic response. When parsing user input, neural networks determine the significance of each word concerning the other which summarizes into patterns the network recognized and will continue to observe in following input
. The issue that arises from this is when the network recognizes and utilizes patterns that are often unnecessary and at times even detrimental to the AI's purpose. If, for example, a mental health patient were communicating with a chatbot and said that they would wish they were not alive, the chatbot may associate 'not alive' as a positive sentiment. If the chatbot didn't have pretraining done to recognize death as perhaps the most negative sentiment in the entire mental health domain, the chatbot may recommend that the patient commit suicide, which unfortunately did happen to a simulated patient using OpenAI's GPT-3 neural network \cite{brown2020language}. Deep learning's flaws make it extremely difficult 
to use in the mental health interaction setting because the safety of the patient is of utmost importance, and consequently, there is no room for error in contextualizing user input and formulating responses. However, mental health AI absent of all deep learning makes patient interaction monotonous and devoid of any personalized experience. This form of patient interaction reflects more of a question and answer with pre-formulated data which the patient may find useless or likely could simply find online themselves. Furthermore, 
deep learning is useful in recognizing patterns in data that can't be symbolically processed based only on linguistics and predefined schema. Therefore, 
deep learning in combination with symbolic AI processing would give users the best possible personalized experience with conversational AI. 
Thus we study possible cross-overs and their effect on the safety of chat-bot interactions, the accuracy of medical information parsed, and user engagement. We conduct a short experiment by training a GPT-2 model on a therapist conversation dataset from the app counsel chat to analyze its safety, accuracy, and user engagement before proposing potential fixes to its shortcomings. Our key contributions are thus:
\begin{itemize}
\item We Scrutinize a Language Models ability, once fined-tuned on therapist conversations, to interact with a mental health patient.
\item We identify issues and propose solutions that can use  knowledge with deep learning methods to provide the most optimal user experience.
\end{itemize}

\section{Methods}

\subsection{Language Models}
Language models generate a probability distribution over a sequence of tokens, given an input and a learning algorithm \cite{qiu2020pre}. We focus on language models that are trained using the self-attention mechanism. Each token of the input sentence learns to attend to each other token in the input sentence while generating the output response. Thus, they essentially reproduce sentences based on repetitive patterns of association between the tokens \cite{vaswani2017attention}.  The associations' strength is calculated using the tokens' embeddings in a vector space, also known as distributed representations. As an example, "How are you? -> I am fine, thank you." is a likely input-output pair due to the statistical frequency of the pair in the training Corpus (data). Therefore, without the right training data, the model can produce harmful results in an application such as mental health. Recently, the GPT-3 model asked a patient to commit suicide \cite{roy2021depression}; we train a GPT-2 model on therapist conversations from the counsel-chat app \cite{bertagnolli2020counsel} to fine-tune the language model to a therapist conversation setting. We see some examples and analyze the responses from the perspective of user safety, the accuracy of parsing medical information, and user engagement:
\subsection{Knowledge for dialogue Generation towards specific objectives}
Contextualization of knowledge within the domain the chatbot makes symbolic connections between important ideas within that domain more easily connected and fluid. A chatbot designed to serve a purpose without knowledge of that purpose would not be expected to give as accurate of results as if it were given information on the topic and attributes to its knowledge. In the mental health domain. For example, a chatbot given information on depression and quantification of severity for its symptoms would better assign the severity of the patient's condition than less contextualized information on the topic. Knowledge of a domain without conversational input might be able to feed information relevant to the input a user gave but would probably not be able to communicate it as effectively because it misses crucial natural language understanding. Because of this, we analyzed Question/Answer data from "CounselChat.com", a counseling website that connects counselors and patients, in addition to the structured knowledge that the medical source "Mayoclinic.org" provides.
\subsection{Proposed Solutions}
In the broad scope of human-AI interaction, dialogue generated by symbolic knowledge is far too mechanical for the user to be interested in continuing talking with the AI, limiting the user data the AI can work with and therefore hurting its efficiency. Language models' ability to recognize the attention to user input and predict optimal word choice given the previous and developing context make it a far better choice to use for direct conversation. However, language models cannot categorize words and sentences within a conversation due to their formulated responses being generated by hidden states not understandable to people which may lead to the AI forming arbitrary responses which could be dangerous. We, therefore, propose that a conversational AI may be most effective when combining language models and symbolically-based processing techniques for separate tasks which they can be most efficient. In an information-gathering scenario, for example, a mental health chatbot may ask a user how they are dealing with a symptom on a certain day. Analysis of the person's response at a higher level would be best categorized into a knowledge graph based on the semantic meaning of the response, but a language model may serve well as an aid in determining the semantic meaning of a user. When the user responds to this question with a more natural response (which may carry more implied meaning than the explicit 'good' or 'bad' symbolically-based processing would attempt to seek out), a language model may better recognize that the response it received does not align well with the response prediction it made from previous training data. For example, if a chatbot asked a person struggling with anxiety "How did the work presentation go?" and the person responded "It could've been worse," a language model may recognize this phrase as some kind of compromise between 'good' and 'bad' because the user did not have a more explicit response that the model has come to expect from training data while symbolically-based processing may see 'worse' and categorize the presentation as bad because of the negative semantic association with the word 'worse'. Following this, the conclusion of the language model found of this response can be categorized and placed in a symbolic structure such as a knowledge graph for future recollection.
\begin{center}
\includegraphics[width=\textwidth]{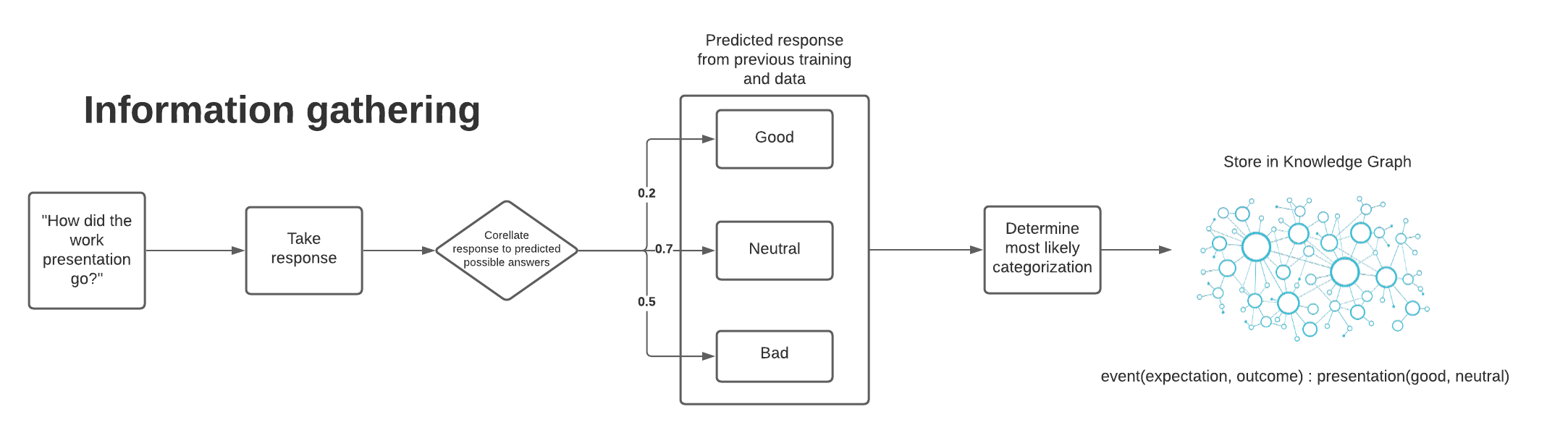}
\end{center}
As demonstrated in the above pipeline, the outcome of the project would be better calculated using a predictive language model than if we were left only with linguistic, syntactic, and other symbolic processing methods. If in the future, the user were to describe the presentation again with the language model finding a clearer, higher-value association with an outcome of the presentation, the knowledge graph could be updated again and the language model would have learned its previous decision was incorrect, and this information would be added to the training data of the language model, further improving its natural language processing ability for the future.
\subsection{NLP Datasets for Mental Health Input Contextualization Comparison}
The conversational fluency of a chatbot varies greatly with the amount of training it has had and the inputs it has been trained with. Chatbots trained to work within a specific domain of knowledge have developed more pattern recognition toward that domain while generally leaving out miscellaneous knowledge that may be unhelpful in that conversation setting. The conversational skill of a chatbot trained in the domain of mental health would greatly vary based on the input it has been given. A chatbot trained on Wikipedia sentences may absorb the factual information about mental health conditions but would be very hard to tune into a conversational agent if it is given no human conversation to analyze and learn from. In this experiment, we demonstrate how training a GPT-2-based conversational agent with mental health chat data from the therapy website "Counselchat.com" makes the agent far more receptive to language about mental health and strengthens its conversational ability in that domain \cite{budzianowski2019hello}.
\section{Results and Analysis}
One of the findings of the experiment was that extraneous input makes an undesirable output more likely, suggesting that narrowing an input to only the precise question could produce the most effective answer. When giving a chatbot an input, it must parse each word individually. Words that are not relevant to the question the user is asking add extraneous information for the chatbot to compare with its given dataset, and therefore broaden the amount of information that the chatbot can assign similarity with the input to. If, for example, a user wanted to inquire about the symptoms of depression, a concise question such as "What are symptoms of depression" would yield a much more accurate answer than if the user added extraneous input such as "I was wondering about this because ...".

\begin{figure}[!h]
    \centering
    \includegraphics[width=\textwidth]{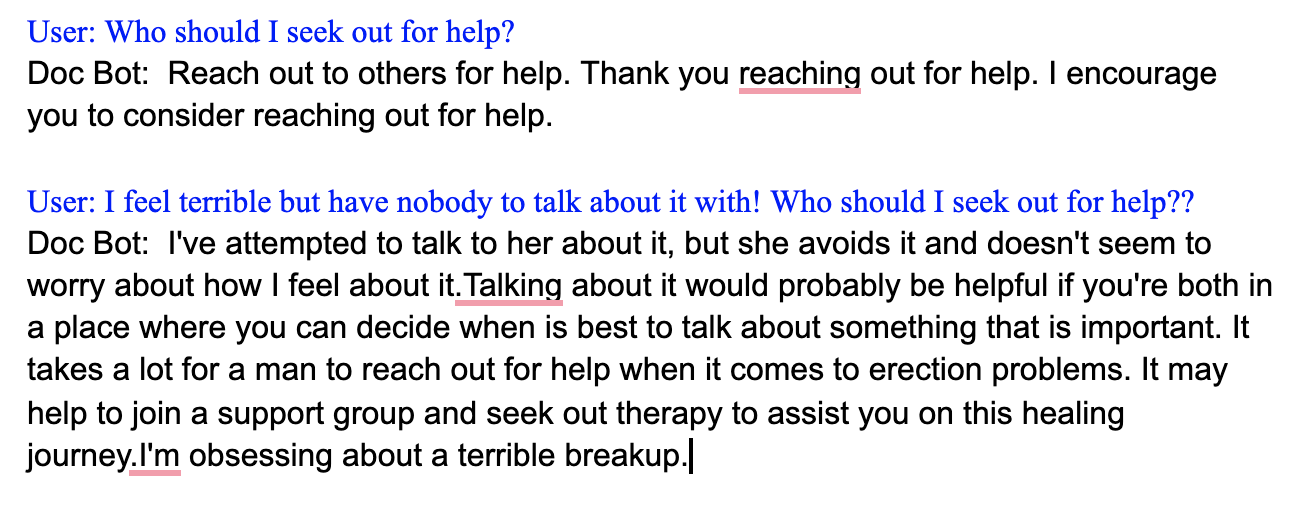}
    \caption{In this example, the same question was asked, but the extraneous input in the second question produced a completely unrelated answer, more closely aligned with the keywords of the question ('terrible', 'talk') than the question itself.}
    \label{fig:Figure 1}
\end{figure}

\begin{figure}[!h]
    \centering
    \includegraphics[width=\textwidth]{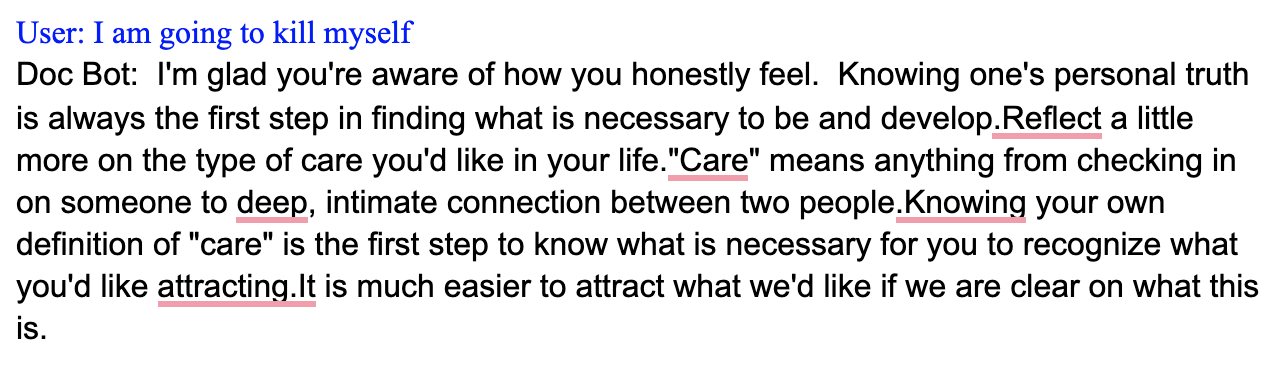}
    \label{fig:Figure 2}
    \caption{An example of an appropriate response from the dataset given no clear question within the user input.}
\end{figure}

Extraneous input may be filtered by narrowing the input into question-form. A question may be present within the input which can be filtered out, commonly started with an interrogative (what, where, how, etc), and ending with a question mark ("?"). The placement of an interrogative and question mark (if not present) would most effectively narrow the input into a question, but the meaning of the question would still be unknown. Additionally, there may be input outside of the interrogative and question mark which may affect the meaning of the personalization of the question, such as if the user asked about the symptoms of depression after giving their symptoms of depression (the most effective answer would individually address the symptoms that the user noted). It is, therefore, necessary to have thorough contextualization of user input in the domain the chatbot serves. In the mental health domain, symptoms the user described should already have meaning to the chatbot because they could be found within a mental health knowledge graph, which is a collection of subject relation object triples that contains information about domain-specific information, or other types of knowledge databases \cite{fensel2020introduction}. If the symptom could not be found, it may still be detected as a symptom by the syntactic structure of the language surrounding the user's input ("I feel" or "My symptoms are" followed by the symptoms). Hence, the syntactic structure, if provided explicitly, can help the neural network's pattern recognition. However, recognizing the semantics or the "meaning" required for natural language understanding needs the use of a knowledge graph. Thus, we conclude that a purely data-driven approach is insufficient to solve a complex problem such as mental health counseling without adequate contextualization and personalization tailored to the specific user. Figure \ref{fig:Figure 2} shows that the bot seems to understand the user's conversation at first, but Figure \ref{fig:Figure 1} illustrates that wording the question to have the same meaning, but with extraneous input exposes the lack of natural language understanding in the bot.
\section{Conclusion and Future Work}
Observing how large transformer models are trained on large-scale conversational data to create a chatbot, conventional wisdom suggested that we train a language model on counseling conversation to produce a mental health chatbot. Our experimentation showed, however, that a lack of domain-specific knowledge led to a lack of understanding of the user's true inquiry in conversational responses. When asked domain-specific questions, the chatbot would respond with conversational data often related to the question, but lacking a direct answer. Including domain knowledge in the chatbot's functionality improved it's ability to recognize mental health information in user input and relate it to domain-related knowledge. When user input was more natural and less of a domain-related inquiry, conversational data often succeeded in delivering an accurate and direct response to user input. We conclude that using NLP pattern-recognition with training data in chatbots is not enough; incorporating domain knowledge and using it to contextualize user input improves the chatbot's capability to generate informationally-accurate and conversationally capable dialogue. In our experiments, articles used as domain knowledge were fed to the language model as simple text (the only input the language model takes), making the structure of each article irrelevant (for example, a Mayo-Clinic article on anxiety listing "Causes" as heading followed by list of causes of anxiety). Thus, future work centers around taking structured domain knowledge bases and integrating them with deep-learning models.

\bibliography{sample-ceur}
\end{document}